\PassOptionsToPackage{table}{xcolor}
\documentclass[10pt,twocolumn,letterpaper]{article}

\usepackage{iccv}              

\usepackage{enumitem}
\usepackage{soul}

\definecolor{iccvblue}{rgb}{0.21,0.49,0.74}
\definecolor{best}{RGB}{8,221,125}
\definecolor{secbest}{RGB}{188,243,215}
\usepackage[pagebackref,breaklinks,colorlinks,allcolors=iccvblue]{hyperref}
\usepackage{multirow}

\usepackage{pifont}

\def\paperID{5317} 
\def\confName{ICCV}
\def\confYear{2025}

\title{Statistical Confidence Rescoring for Robust 3D Scene Graph Generation from Multi-View Images}

\author{Qi Xun Yeo \quad Yanyan Li \quad Gim Hee Lee  \\
Department of Computer Science, National University of Singapore\\
{\tt\small \{qixunyeo,yan.li,gimhee.lee\}@nus.edu.sg}
}

\begin{document}
\maketitle
\begin{abstract}

Modern 3D semantic scene graph estimation methods utilize ground truth 3D annotations to accurately predict target objects, predicates, and relationships. In the absence of given 3D ground truth representations, we explore leveraging only multi-view RGB images to tackle this task. To attain robust features for accurate scene graph estimation, we must overcome the noisy reconstructed pseudo point-based geometry from predicted depth maps and reduce the amount of background noise present in multi-view image features. The key is to enrich node and edge features with accurate semantic and spatial information and through neighboring relations. We obtain semantic masks to guide feature aggregation to filter background features and design a novel method to incorporate neighboring node information to aid robustness of our scene graph estimates. Furthermore, we leverage on explicit statistical priors calculated from the training summary statistics to refine node and edge predictions based on their one-hop neighborhood. Our experiments show that our method outperforms current methods purely using multi-view images as the initial input. Our project page is available at \url{https://qixun1.github.io/projects/SCRSSG}.
\end{abstract}    
\section{Introduction}
\label{sec:intro}

The semantic scene graph (SSG) is a crucial intermediate representation that enhances higher-level scene understanding. It plays a key role in tasks such as image captioning \cite{yang2019auto, xu2019scene, nguyen2021defense}, image retrieval \cite{schuster2015generating, johnson2015image, schroeder2020structured, yoon2021image}, image editing \cite{chen2020graph, dhamo2020semantic, zhang2023complex}, and medical applications \cite{lin2022sgt, d2023knowledge, holm2023dynamic, ozsoy20224d, yuan2024advancing} by capturing both semantic and, more importantly, relational information between objects and their surroundings.
Initially developed for 2D images, SSG has since expanded to the 3D domain \cite{dhamo2021graph, ost2021neural, gao2024graphdreamer}. 3D scene graphs provide a high-level representation of an entire 3D scene using inputs such as multi-view RGB images or LiDAR point clouds. Unlike their 2D counterparts, they incorporate spatial relationships that extend beyond the visible image plane, allowing a holistic understanding of complex environments.

\begin{figure}[t]
  \centering
   \includegraphics[width=\linewidth]{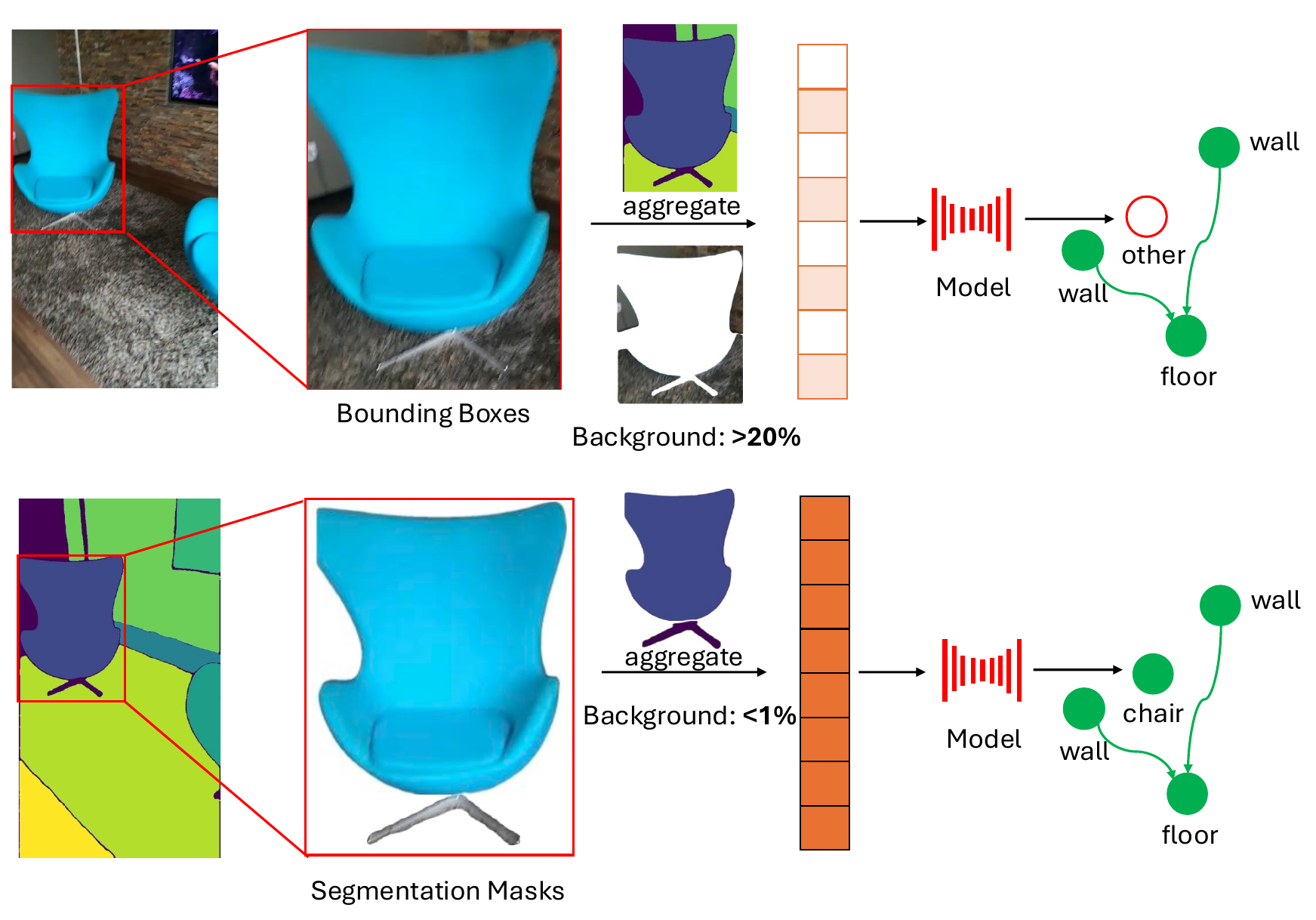}
   \vspace{-0.1in}
   \caption{
   Existing multi-view RGB methods aggregate features using bounding boxes, introducing background noise that hinders accuracy. We use pretrained segmentation masks reduce this interference and thus improving prediction accuracy.}
   
   \label{fig:teaser}
   \vspace{-0.1in}
\end{figure}

Many approaches \cite{armeni20193d, wald2020learning, wald2022learning} rely on ground-truth 3D LiDAR point clouds as input. However, LiDAR sensors are resource-intensive and lack inherent semantic richness. 
Recent works \cite{wu2021scenegraphfusion, wu2023incremental} have shifted toward using semantically rich multi-view RGB images, which offer a balance between computational efficiency and improved semantic fidelity.
Although point cloud-based methods remain valuable for capturing geometric information, our approach focuses on leveraging multi-view RGB images. By incorporating pretrained depth estimators, we generate pseudo point-based geometry to estimate the 3D semantic scene graph, reducing reliance on LiDAR while preserving spatial and structural details.

A key challenge in using only multi-view RGB images is ensuring that node features remain free from distractors. As shown in Fig. \ref{fig:teaser}, prior approaches that rely on entity detectors to extract 2D bounding box proposals or regions of interest often fail to guarantee robustness since distractors within the bounding box are often mistakenly aggregated. Ensuring robust initial multi-view features is crucial without ground truth point-based geometry for error correction. No explicit mechanism exists to correct incorrectly bounded objects or misclassified rare classes that are not well trained on the model. To alleviate this problem, a mechanism should exist to refine estimates using prior knowledge instead of depending solely on implicit interactions between nodes and edges to achieve confident and accurate predictions.

In this paper, we propose a framework to enhance the robustness of the model. First, we introduce a \textbf{\textit{masked feature initialization}} (MFI), which leverages segmentation masks to aggregate image features instead of relying on the bounding box proposals. This reduces background noise, which results in cleaner multi-view image features.
Next, we design a robust \textbf{\textit{residual spatial neighbor graph neural network}} (RSN-GNN) to encode spatial information into node features. This network filters highly activated regions from neighboring nodes, refining target node features for improved predicate estimation.
Finally, we propose a \textbf{\textit{confidence rescoring}} (CR) module, which refines object and predicate estimates using an inverse softmax-weighted contribution of neighboring node-to-node and node-to-edge co-occurrence counts. By integrating this explicit inductive bias, our approach improves the accuracy of the prediction, particularly in low-confidence scenarios. Extensive experimental results on the benchmark 3RScan dataset show the competitiveness of our proposed approach compared to existing 2D and 3D approaches. 

\noindent Our main contributions can be summarized as follows:
\begin{itemize}[leftmargin=0.5cm]
    \vspace{0.8mm}
    \item We introduce a masked feature initialization to enhance the robustness of node features by reducing background distractors, yielding cleaner multi-view image features.
    \vspace{0.8mm}
    \item We design a novel GCN architecture that integrates highly activated neighboring features into the target node to increase the robustness of the edge features.
    \vspace{0.8mm}
    \item We propose a new refinement module to explicitly refine predictions based on statistical prior knowledge.
    \vspace{0.8mm}
    \item Experiments show that we outperform previous state-of-the-art approaches on the 3RScan dataset, particularly on metrics that deal with low-tail imbalanced classes.
\end{itemize}

\section{Related Work}
\label{sec:related}

\noindent \textbf{2D Semantic Scene Graph.} 
2D semantic scene graph prediction is typically categorized into two-step and one-step approaches. The two-step approach first detects objects and then classifies their relationships \cite{Yang_2018_ECCV, baier2017improving, zellers2018neural, chen2019knowledge}. In contrast, the one-step approach jointly infers object and relationship classes \cite{xu2017scene}.
%
Xu et al. \cite{xu2017scene} pioneers the problem of scene graph generation and they tried to solve this problem via iterative message passing. Baier et al. \cite{baier2017improving} first showed how semantic models can be improved by incorporating triplet frequencies. Zellers et al. \cite{zellers2018neural} analyzes the usefulness of statistical co-occurrences for the Visual Genome dataset and concluded that such statistical priors serves as strong regularization for the task. Chen et al. \cite{chen2019knowledge} formalizes the first approach (KERN) to incorporate the statistical prior directly into graph neural networks. Sharifzadeh et. al. proposes Schemata to assimilate image-based relational prior knowledge into the representations within the neural network \cite{sharifzadeh2021classification}. Compared to previous methods, our method combines predictions from the prediction head with statistical co-occurrence of the current node and neighbors to determine the final class using a confidence score. Unlike Sonogashira et. al. \cite{DBLP:journals/access/SonogashiraIK22} which uses hard thresholding
, our approach uses adaptive relationship and node refinements.

\vspace{2mm}
\noindent \textbf{3D Semantic Scene Graph.} 
Methods for generating a 3D semantic scene graph can be categorized into two types based on input data. The first relies on 3D inputs such as ground-truth 3D geometry to learn contextual information \cite{wald2020learning, zhang2021exploiting, wang2023vl}. The second relies on multi-view RGB images with some methods incorporating depth information \cite{wu2021scenegraphfusion} while others do not \cite{wu2023incremental}. Our approach follows the latter, leveraging the estimated depth from off-the-shelf depth estimators \cite{yang2024depth} to enhance the information provided by multi-view RGB images.
Wald et al. \cite{wald2020learning} first proposed 3RScan: a richly annotated 3D scene graph dataset and a method that focuses on pairwise relationships to predict the 3D scene graphs. Wu et al. \cite{wu2021scenegraphfusion} uses SLAM to obtain dense point representations and a novel graph convolutional network (GCN) based aggregation function to enhance 3D scene graph prediction. Zhang et al. \cite{zhang2021exploiting} proposes an edge-oriented GCN to incorporate multi-dimensional edge features for explicit inter-node relationship modeling. Zhang et al. \cite{zhang2021knowledge} integrates prior commonsense knowledge by learning meta-embeddings only from the class labels with their graphical structures to avoid perceptual errors. 
Wang et al. \cite{wang2023vl} incorporated additional information by distilling knowledge learned by the multi-modal oracle model to the 3D model. Wu et al. \cite{wu2023incremental} introduces a novel entity association method to obtain the 3D entities from the predicted 2D entities and a geometric gate to incorporate geometric information to multi-view image features. Feng et al. \cite{feng20233d} uses ConceptNet external knowledge base to accumulate both contextualized visual content and textual facts to form a 3D spatial multimodal knowledge graph. Our method leverages on the explicit statistical priors similar to 2D scene graph method KERN and Schemata while guiding low confidence predictions made by the initial node and edge predictors using both visual and geometric cues.

\section{Our Methodology}
\label{sec:methodology}

\begin{figure*}[t]
  \centering
  \includegraphics[width=\textwidth]{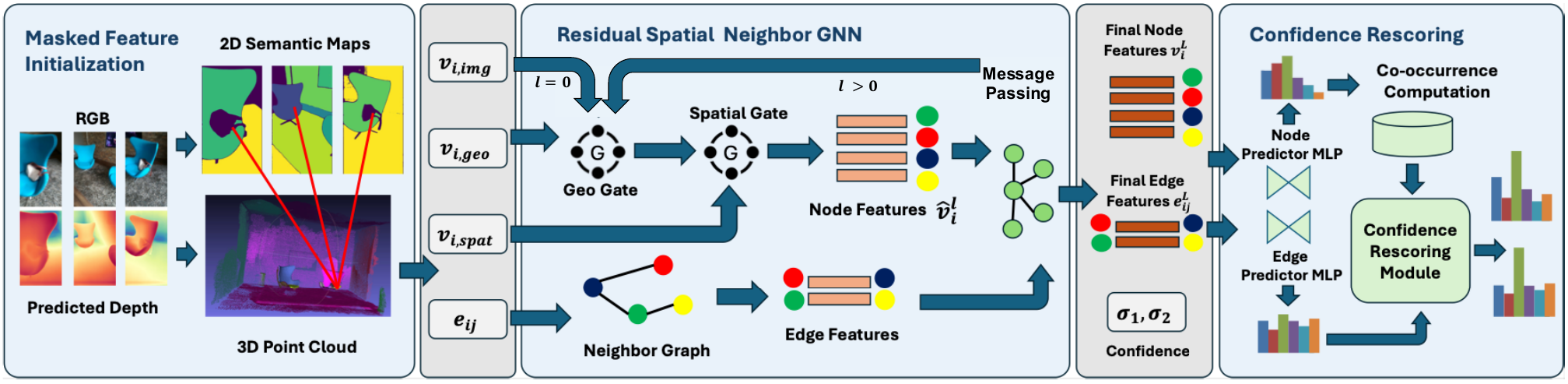}
  \caption{
  The architecture of our proposed framework consists of three main components: 1) A masked feature initialization step leverages backprojected semantic masks to aggregate multi-view image, geometric, and spatial features. 2) A message passing step utilizes our residual spatial neighbor GNN module to refine node and edge features through feature interaction. 3) A confidence rescoring step adjusts logits based on summary statistics of object and predicate co-occurrence from the training dataset.}
  
  \label{fig:architecture}
\end{figure*}

\noindent \textbf{Problem Definition.}
Given either a set of 3D point clouds $P$ or multi-view RGB images $I$ of a 3D scene as input, the goal is to estimate the 3D scene graph denoted as:
\begin{equation}
    G = (O, R), 
\end{equation}
where 
\begin{itemize}[leftmargin=0.5cm]
    \item $N$: Number of multi-view input images;
    \item $O = \{o_{m}\}_{m=1}^{M}$: All objects in the scene;
    \item $R = \{r_{k \rightarrow m}\}_{m,k=1}^{M,M}$: Relationships between all objects;
    \item $M = \lvert O \rvert$: Number of objects in the scene.
\end{itemize}
\vspace{1mm}
Furthermore, we refer to $o_{m}$ as the object class for node instance $m$, $r_{k \rightarrow m}$ as the predicate class for the edge connecting nodes $k$ and $m$. Alternatively, the 3D scene graph can be defined as a set of triplets given by: $\{o_{k}, r_{k \rightarrow m}, o_{m}\} := \{\operatorname{node}, \operatorname{predicate}, \operatorname{node}\}$.


\vspace{2mm}
\noindent \textbf{Overview.}
Fig.~\ref{fig:architecture} shows an illustration of the architecture of our proposed framework. The input to our framework is a set of $N$ multi-view RGB images $\{I_i\}_{i=1}^{N}$ of the 3D scene. Since the focus of our work is not on multi-view 3D reconstruction, we recover the 3D point cloud of the scene by running an off-the-shelf RGB-D SLAM system \cite{tateno2015real} with the metric depth maps of the input images $\{\hat{D_i}\}_{i=1}^{N}$ obtained from the pretrained Depth Anything \cite{yang2024depth} model. 

From the predicted 3D point cloud, we initialize multi-view image features using segmentation masks, extract geometric features with a point-based feature extractor, and derive spatial features via a linear layer. These multi-view image features serve as the initial input to our RSN-GNN module.  
At the GCN initialization, we enhance node features by integrating geometric and spatial features through geometric and spatial gates. We additionally enhance edge features by integrating max pooled neighboring nodes features. These enhanced node and edge features undergo message passing, with the integration process repeating at each layer. After \( k \) iterations, the final node and edge features are processed by predictors to generate initial estimates, which are then refined using precomputed co-occurrence statistics via our CR module, producing the final scene graph output.

We elaborate on the feature initialization step in Sec.~\ref{sec:feat}, the message passing step in Sec.~\ref{sec:message}, and the confidence rescoring step in Sec.~\ref{sec:rescore}.

\subsection{Masked Feature Initialization (MFI)}
\label{sec:feat}
We initialize the node features of the scene graph by combining multi-view image representations from a pretrained image encoder with PointNet features derived from the predicted points in the RGB-D SLAM reconstruction step that utilizes predicted depth. To eliminate background clutter, we employ the Segment Anything Model (SAM)~\cite{kirillov2023segment} to produce precise 2D semantic masks for each object. For each node, we backproject the associated points onto the 2D instance mask to identify its corresponding region. We then generate object proposals from these instance masks, resize them to $224\times224$, and extract their multi-view image features using the pretrained encoder. Finally, we apply these semantic masks to the multi-view image features before aggregating them into our node representations.

Our method to get the object proposals, \ie the nodes of the scene graph offers two key advantages over relying solely on the entity detector from SceneGraphFusion and JointSSG~\cite{wu2021scenegraphfusion, wu2023incremental}. First, the entity detector often captures only part of an object. Although SceneGraphFusion propagates the label for this segment, the feature extractor can operate only on the detected portion. In contrast, our method employs more complete object masks, enabling the feature extractor to derive features from the entire object. Second, we exclude extraneous background features by applying these masks before feature aggregation. In SceneGraphFusion and JointSSG, the bounding boxes of the pre-trained entity detector frequently include background regions, which can dilute the quality of the features and hinder the performance. Our method is closer to the open vocabulary method ConceptGraphs~\cite{gu2024conceptgraphs} which uses SAM masks but with the aggregation method provided by JointSSG.

Formally, we compute a corresponding multi-view image feature \( \mathbf{v}_{i, img} \), a geometric feature \( \mathbf{v}_{i, geo} \), and a spatial feature \( \mathbf{v}_{i, spat} \) for each node \( v_i \). We aggregate features only for nodes that are visible in the covisibility graph\footnote{The covisibility graph is obtained as intermediate output after running the RGB-D SLAM system. It refers to a bipartite graph where each image \( I_k \)is connected with the specific node instance \( v_i \).} \( G_v \) denoted by \(\mathbf{1}(v_i \in I_k) \) and have corresponding pixels within the semantic mask provided by SAM. The aggregated multi-view image feature \( \mathbf{v}_{i, img} \) is given by:
\begin{equation}
\label{eq:imageFeatures}
\mathbf{v}_i^0 = \frac{\sum_{k=1}^{N}   \mathbf{1}(v_i \in I_k)f(\eta_{k} \odot x_{i,k})}{\lvert k \in G_v \rvert},
\end{equation}
which also serves as the initial node feature. $f(\cdot)$ is the image feature encoder and $\eta_k$ refers to the 2D semantic mask. $x_{i,k}$ refers to the cropped image patch containing the node \( i \) from the $k$-th image. $\odot$ refers to the element-wise multiplication operator. The aggregation is the arithmetic mean of the node features for each image in the covisibility graph. 

The geometric feature \( \mathbf{v}_{i, geo} \) is extracted using a vanilla PointNet~\cite{qi2017pointnet}, with inputs derived from the 3D instance masks generated during the RGB-D SLAM step. For the spatial features, we employ a simple linear layer to transform the attributes of each 3D bounding box: dimensions \( \mathbf{b}_i \), volume \( s_i \), length \( l_i \), and the ratio of its x- and y-axes \( r_i \) into the spatial feature as: 
\begin{equation}
\label{eq:spatialFeature}
\mathbf{v}_{i, spat} = g_s([\mathbf{b}_i, s_i, l_i, r_i ]),
\end{equation}
where \( g_s \) refers to a set of learnable weights for spatial features. \( [\cdot] \) refers to the concatenation of the inputs within the operator.

\subsection{Residual Spatial Neighbor GNN 
Module}
\label{sec:message}
Following a similar design to the learnable geometric gate in JointSSG~\cite{wu2023incremental}, we integrate both geometric and spatial features into the initial node feature, which originally consists only of multi-view image features. The spatial features serve as additional information for individual nodes, analogous to the embeddings used for token encoding in Transformer architectures~\cite{vaswani2017attention}. We define the geometric gate as:
\begin{equation}
\label{eq:geoGate}
\hat{\mathbf{v}}_i^{l} = \mathbf{v}_i^{l} + \sigma(\mathbf{w}_{g}^T[\mathbf{v}_i^{l}, \mathbf{v}_{i, geo}]) \cdot \sigma(\mathbf{v}_{i, geo}),
\end{equation}
where $\mathbf{w}_g$ refers to a set of learnable weights, $\sigma$ refers to the sigmoid operator and $l$ refers to the layer of the node feature. Furthermore, we define the spatial gate as:
\begin{equation}
\label{eq:spatialGate}
\ddot{\mathbf{v}}_i^{l} = \hat{\mathbf{v}}_i^{l} + \sigma(\mathbf{w}_{s}^T[\hat{\mathbf{v}}_i^{l}, \mathbf{v}_{i, spat}]) \cdot \sigma(\mathbf{v}_{i, spat}),
\end{equation}
where $\mathbf{w}_s$ is a set of learnable weights for spatial features.






Inspired by the residual connections in ResNet~\cite{he2016deep}, we introduce a mechanism to integrate highly activated neighboring features into the target node feature using max pooling during the message-passing process for edge features. This approach implicitly encodes neighboring information, improving edge prediction for scene graph estimation.
Specifically, we augment the edge messages in the GCN by integrating max-pooled neighboring features before message passing as follows:
\begin{equation}
\label{eq:maxPoolMsgPass}
\tilde{\mathbf{v}}_i^{l} = \ddot{\mathbf{v}}_i^{l} + \max([\ddot{\mathbf{v}}_j^{l}]_{j\in n(i)}),
\end{equation}
where $\max(\cdot)$ refers to the max pooling operator and $n(i)$ refers to the neighboring nodes of node $i$. Max pooling allows highly activated regions in the different neighboring node features to be fused into the target node feature via the residual connection.





We apply max-pooled neighboring features to the calculation of edge messages without using them in node message updates. Since node features are highly susceptible to noise and spurious correlations from other nodes~\cite{lin2022ru}, incorporating max-pooled neighboring features into node message computations can introduce confusion instead of improving performance. The edge message \( m_{i \rightarrow j} \) is thus defined as:
\begin{equation}
\label{eq:edgeMsg}
m_{i \rightarrow j} = g_e([\tilde{\mathbf{v}}_i^{l}, \mathbf{e}_{ij}^{l}, \tilde{\mathbf{v}}_j^{l}]),
\end{equation}
where $g_e(\cdot)$ is an MLP and $\mathbf{e}_{ij}^{l}$ is the edge descriptor used in \cite{wu2023incremental} containing relative node properties such as centroid displacement, relative bounding box size difference, and a relative pose descriptor.




\subsection{Confidence Rescoring (CR) Module}
\label{sec:rescore}

\begin{figure}[t]
  \centering
   \includegraphics[width=0.9\linewidth]{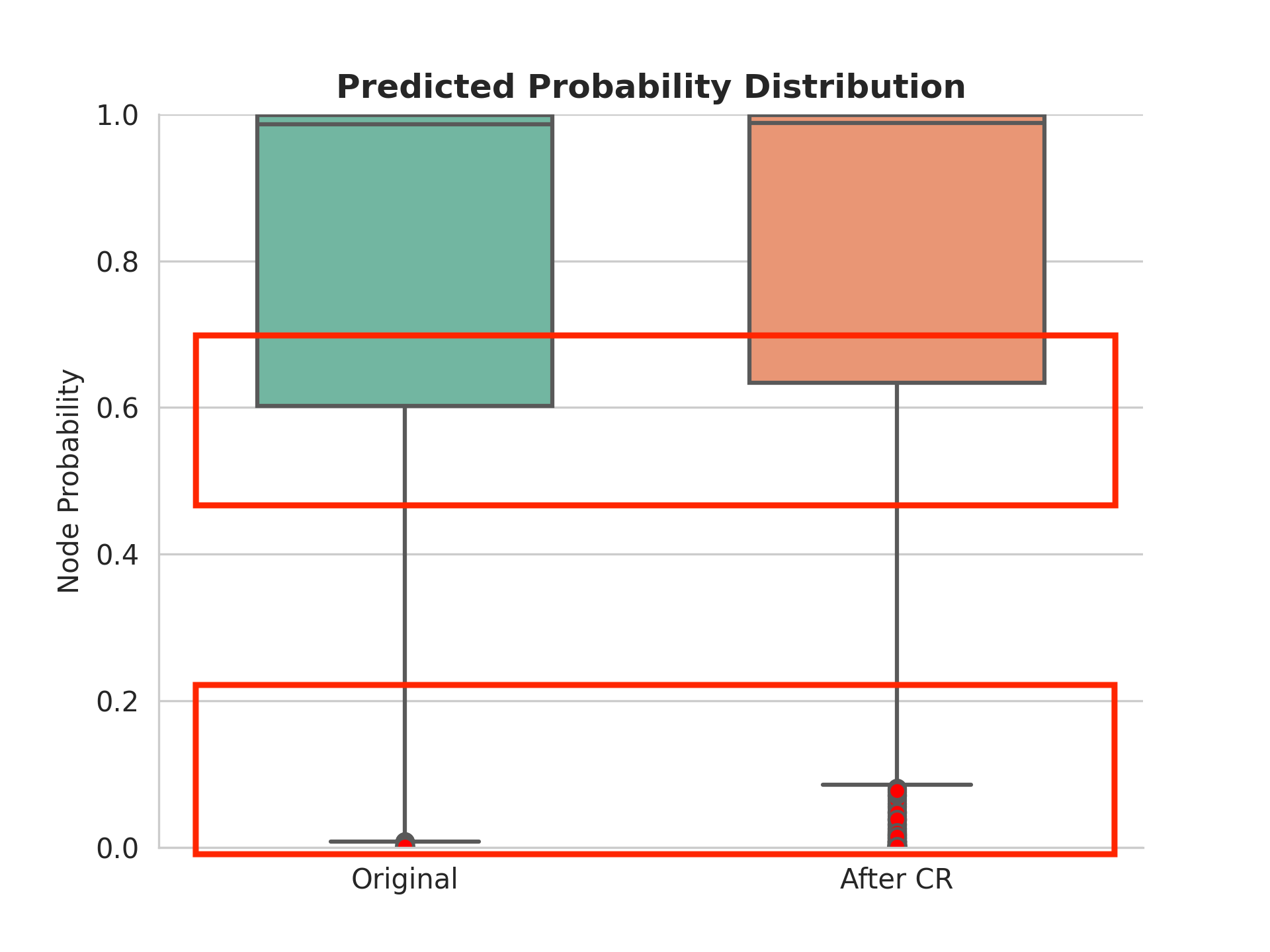}
   \vspace{-0.1in}

   \caption{Box plot for predicted node probability for correct class after our CR module. The lower quartile visibly moves upwards due to an increase in predicted node probability for correct class.}
   \label{fig:logits}
   \vspace{-0.1in}
\end{figure}

As discussed in Sec.~\ref{sec:intro}, modern scene graph methods depend heavily on the entity detector and image feature extractor to produce an accurate initial object prediction. However, there is no mechanism to correct a target bounding box prediction that is initially wrong. To address this limitation, we leverage relationships with neighboring nodes, allowing us to mitigate the impact of incorrect initial predictions by upweighting contributions from neighboring node statistics to refine target object classification. Fig.~\ref{fig:logits} illustrates the logits of the final prediction after applying our confidence rescoring module. This module incorporates prior knowledge to improve prediction confidence beyond that of the original node and edge multi-layer perceptron (MLP) predictor.

The probability of predicting node \( i \) given its neighboring nodes $n(i)$ is approximated by the output of our node predictor \( h_v(\cdot) \) as:
\begin{equation}
\label{eq:orinodeProb}
\begin{split}
    P\left(o_i \mid \{o_j : j \in n(i)\}\right) &= \frac{1}{Z_i} \, \psi_i(o_i) \prod_{j \in n(i)} \psi_{ij}(o_i, o_j) \\
    &\approx \tau(h_v(\mathbf{v}^{L})),
\end{split}
\end{equation}
where $\psi_i(o_i)$ refers to the potential associated with node $i$ and $\psi_{ij}(o_i, o_j)$ refers to the potential associated with the edge between node $i$ and its neighbor node $j$. The potentials are implicitly learned through the GCN unlike classical methods, which explicitly learn these potentials. $h_v(\cdot)$ is a small MLP that predicts the class of each node and $\tau$ refers to the softmax operator. $L$ refers to the last layer of the GCN message passing output. The partition function given by:
\begin{equation}
\label{eq:nodePartitionFunc}
    Z_i = \sum_{o_i} \psi_i(o_i) \prod_{j \in n(i)} \psi_{ij}(o_i, o_j)
\end{equation}
ensures the probabilities sum to 1. A similar equation can be obtained for the probability of predicting edge $ij$ given its neighboring nodes $i$ and $j$, \ie:
\begin{equation}
\label{eq:edgeProb}
\begin{split}
    P\left(r_{i \rightarrow j} \mid o_i, o_j\right) &= \frac{1}{Z_{ij}} \psi_{ij}(o_i, o_j, r_{i \rightarrow j}) \\
    &\approx \tau(h_e(\mathbf{e}^{L})),
\end{split}
\end{equation}
where $h_e(\cdot)$ is a small MLP which predicts the class of each edge, and the partition function to normalize the probabilities is given by:
\begin{equation}
\label{eq:edgePartitionFunc}
    Z_{ij} = \sum_{r_{i \rightarrow j}} \psi_{ij}(o_i, o_j, r_{i \rightarrow j}).
\end{equation}
%
%
%
%
%
%
%
We begin by computing summary statistics from the training dataset to node-to-node co-occurrences denoted as \( c_{ij}(o_i, o_j) \), and node-to-edge co-occurrences denoted as \( d_{ij}(o_i, e_j) \). Subsequently, we determine the prediction confidence of the model, where \( \alpha_{\mathbf{v}} \) represents node confidence and \( \alpha_{\mathbf{e}} \) represents edge confidence.
Next, we obtain the node $\alpha_{\mathbf{v}}$ and edge $\alpha_{\mathbf{e}}$ confidences as:
\begin{equation}
\alpha_{\mathbf{v}} = \max_{\mathbf{v}} \tau(h_v(\mathbf{v}^{L})), \quad
\alpha_{\mathbf{e}} = \max_{\mathbf{e}} \tau(h_e(\mathbf{e}^{L})). 
\end{equation}
We first estimate the marginal probability of a specific instantiation of node \( i \), conditioned on node \( j \) using node-to-node co-occurrences, and on edge \( ij \) using node-to-edge co-occurrences defined as:
\begin{equation}
\begin{split}
\label{eq:condProb}
    & \mathbf{c}(o_{i}| o_{j}) = [c(o_i=c_k| o_j = c_l)]_{k=0,l=0}^{C, C}, \quad \text{where} \\
    & c(o_{i} = c_k| o_{j}=c_l) = \frac{c(o_i=c_k, o_j = c_l)}{\sum_{k=0}^C c(o_i=c_k, o_j = c_l)}.
\end{split}
\end{equation}
Here, \( C \) refers to the number of object classes and \( c_k \) refers to the \( k \)-th class in the set of object classes. We then integrate the summary statistics via an inverse softmax operation denoted as \( \gamma(\cdot) \) into the node probability in Eq.~\ref{eq:nodeProb} to get:
\begin{equation}
\label{eq:nodeProb}
\begin{split}
    &P\left(o_i \mid \{o_j : j \in n(i)\}\right) = \tau(\alpha_{\mathbf{v}} \cdot h_v(\mathbf{v}_i) \\ & \qquad \qquad + (1-\alpha_{\mathbf{v}}) \cdot \sum_{j\in n(i)} \alpha_j \cdot \gamma(\mathbf{c}(o_i| o_j))).
\end{split}
\end{equation}
This formulation is chosen because directly adding the estimated marginal probability to the predictions of the model after softmax would disregard the required normalization, resulting in incompatible terms.

%
%

%
%

%

Similarly, we compute the estimated marginal probability of an edge \( ij \) instantiation conditioned on nodes \( i \) and \( j \) using node-to-edge co-occurrence statistics for predicate estimation: 
\begin{equation}
\label{eq:predicateEst}
\begin{split}
    & \mathbf{d}(r_{i \rightarrow j} \mid o_{j}) = [d(r_{i \rightarrow j}=e_k| o_j = c_l)]_{k=0,l=0}^{P, C}, ~~ \text{where} \\
    & d(r_{i \rightarrow j=k} \mid o_{j=l}) = \frac{d(r_{i \rightarrow j}=e_k, o_j = c_l)}{\sum_{k=0}^P d(r_{i \rightarrow j}=e_k, o_j = c_l)}.
\end{split}
\end{equation}
Here, \( P \) refers to the total number of predicate classes and \( e_k \) refers to the \( k \)-th class in the set of predicate class.
The refined logits for predicate estimation are then given by:
\begin{equation}
\label{eq:logitPredicates}
\begin{split}
    P\left(r_{i \rightarrow j} \mid o_i, o_j\right) &= \tau(\alpha_{ij} \cdot h_e(\mathbf{e}_{ij}) \\ &+ (1-\alpha_{ij}) \cdot  (\alpha_i \cdot \gamma(\mathbf{d}(r_{i \rightarrow j}| o_{i}))) \\ &\cdot(\alpha_j \cdot \gamma(\mathbf{d}(r_{i \rightarrow j}| o_{j})))).
\end{split}
\end{equation}
\noindent \textbf{Remarks.} As shown in Fig.~\ref{fig:logits}, the original node probability for the correct class is already relatively high. However, our CR module further enhances the predicted probability as evidenced by the upward shift in the box plot from ``Original" to ``After CR" in the red boxes, with a notable increase in the lower quartile.
Moreover, a larger proportion of low node probability instances are flagged as outliers as shown by the greater number of red circles at ``After CR" compared to ``Original". 
Most notable gains occur when original logits are low since CR increases the probability of the correct class by down-weighting misclassified classes $c^\prime$ via lower conditional probabilities $P(c^\prime|c)$ given conditioned class \( c \). This targeted adjustment raises suppressed logits for rare classes, thus improving prediction performance for under-represented categories.


\section{Experiments}
\label{sec:Experiments}

\subsection{Experimental Setup}

\paragraph{Datasets.} 
We evaluate on the 3RScan dataset following the same training and test split from \cite{wald2020learning}. 
The image quality of the dataset is rather poor as each scene constitutes a variable number of short videos spliced together with different motion trajectories. As the images are obtained from consumer-level cameras, the images possess a low frame rate of 10 Hz, image blur, and image jitter from sudden camera motion which makes this dataset challenging.

\begin{table*}[]
    \centering
    \scalebox{0.60}{
    \begin{tabular}{c|cccccccccccccccccccc|c}
    \toprule
      Method   & bath. & bed & bkshf & cab. & chair & cntr. & curt. & desk & door & floor & ofurn & pic. & refri. & show. & sink & sofa & table & toil. & wall & wind. & Mean. \\
      \hline
       IMP  & 0.000 & \cellcolor{best} \textbf{1.000} & 0.000 & 0.341 & 0.467 & 0.000 & \cellcolor{best} \textbf{0.606} & 0.200 & \cellcolor{best} \textbf{0.528} & 0.900 & 0.125 & 0.143 & 0.000 & 0.000 & 0.333 & 0.450 & 0.525 & \cellcolor{best} \textbf{0.800} & 0.622 & 0.125 & 0.358 \\
       VGfM & 0.500 & 0.000 & 0.000 & 0.415 & 0.543 & 0.083 & 0.545 & 0.000 & 0.389 & \cellcolor{best} \textbf{0.920} & 0.232 & 0.086 & 0.000 & 0.000 & 0.381 & 0.450 & 0.574 & \cellcolor{best} \textbf{0.800} & 0.644 & 0.167 & 0.336 \\
       3DSSG & 0.000 & 0.667 & 0.000 & 0.207 & 0.348 & 0.167 & 0.576 & 0.200 & 0.361 & 0.780 & 0.125 & 0.057 & 0.000 & 0.000 & 0.333 & 0.500 & 0.279 & 0.000 & 0.467 & 0.125 & 0.260 \\
       SGFN & 0.500 & 0.333 & 0.000 & 0.293 & \cellcolor{best} \textbf{0.685} & 0.250 & 0.515 & 0.200 & 0.444 & 0.880 & 0.232 & 0.171 & 0.000 & 0.000 & \cellcolor{best} \textbf{0.429} & 0.500 & 0.508 & 0.400 & \cellcolor{best} \textbf{0.707} & 0.125 & 0.359 \\
       JointSSG & \cellcolor{best} \textbf{1.000} & \cellcolor{best} \textbf{1.000} & 0.000 & 0.476 & 0.674 & 0.250 & 0.576 & 0.200 & 0.500 & 0.900 & 0.107 & 0.143 & 0.000 & 0.000 & \cellcolor{best} \textbf{0.429} & 0.550 & 0.541 & \cellcolor{best} \textbf{0.800} & 0.680 & 0.250 & 0.454 \\
       \hline
       Ours & \cellcolor{best} \textbf{1.000}  & \cellcolor{best} \textbf{1.000} & \cellcolor{best} \textbf{0.333} & \cellcolor{best} \textbf{0.622} & 0.630 & \cellcolor{best} \textbf{0.333} & 0.545 & \cellcolor{best} \textbf{0.600} & \cellcolor{best} \textbf{0.528} & 0.880 & \cellcolor{best} \textbf{0.304} & \cellcolor{best} \textbf{0.171} & \cellcolor{best} \textbf{0.667} & \cellcolor{best} \textbf{0.667} & 0.333 & \cellcolor{best} \textbf{0.600} & \cellcolor{best} \textbf{0.623} & \cellcolor{best} \textbf{0.800} & 0.693 & \cellcolor{best} \textbf{0.375} & \cellcolor{best} \textbf{0.605} \\
    \bottomrule     
    \end{tabular}}
    \caption{Comparison with state-of-the-art methods on the 3RScan dataset for each object class. The \colorbox{best}{Best} results are highlighted.}
    \label{tab:perclass}
    \vspace{-0.1in}
\end{table*}

\vspace{2mm}
\noindent \textbf{Implementation Details.}
We utilize the codebase provided from the 3DSSG GitHub repository\footnote{https://github.com/ShunChengWu/3DSSG} to reproduce the prior methods. 
For JointSSG$\ddagger$ \cite{wu2023incremental}, we replace their image feature extractor from a pretrained ResNet18 to a pretrained DINOv2 with a ViT-L backbone to ensure fair comparison.

Following~\cite{wu2023incremental}, we use PointNet~\cite{qi2017pointnet} as the point encoder
and employ two message-passing layers. For RGB-D SLAM reconstruction, we utilize the method from~\cite{tateno2015real} and our predicted depth to generate predicted points. Multi-view feature extraction is performed using a pretrained DINOv2~\cite{oquab2024dinov2} with a ViT-L backbone. To obtain object segmentation masks, we use a pretrained SAM predictor with a text prompt containing the class set of 20 NYU object classes following the subset defined in~\cite{wald2020learning}. Node-to-node and node-to-edge co-occurrences are precomputed on the training set. We train the model on a single RTX 3090 GPU for two days with early stopping.

\vspace{2mm}
\noindent \textbf{Evaluation Metrics.}
Following \cite{wald2022learning} and JointSSG\cite{wu2023incremental}, we report the overall \textbf{top-1} recall (Recall) for the object class estimation (Obj.), the predicate estimation (Pred.), and the relationship triplet estimation (Rel.). We also report the mean recall (mRecall) for the object class estimation (Obj.), the predicate estimation (Pred.) only. We map all predictions on the estimated segments to the ground truth segments for a fair comparison between the different methods.

\subsection{Quantitative Results}
We present the per-class recall for 20 object classes in the 3RScan dataset in Tab.~\ref{tab:perclass}, demonstrating that our method significantly outperforms others across most object categories. Notably, we are the only approach that achieves a recall above 0\% for all object classes, indicating that we \textbf{robustly learn all classes} instead of just a subset. 
Our method tends to underperform slightly compared to the other methods on majority classes such as \( floor \) and \( wall \) classes in return, although not by a significant margin.
The main experimental results for the top-1 recall and the mean recall are shown in Tab.~\ref{tab:main}. 
Importantly, our approach significantly outperforms others in mean recall (mRecall) for both objects and predicates, suggesting that it handles class imbalance issues~\cite{wald2022learning} more effectively than competing methods. 
Additionally, our method surpasses all baselines in relationship and object estimation metrics while achieving comparable performance in predicate estimation. 
More impressively, our method with the weaker ResNet18 multi-view image features already surpasses all baselines in relationship and object estimation metrics except for predicate estimation. Our method with ResNet18 image features even manages to surpass the performance of JointSSG with DINOv2 multi-view image features by non-trivial margins. This shows that our improved performance is not purely 
due to the more powerful DINOv2 features.
Refer to our supplementary material for more quantitative results.


\begin{table}[]
    \centering
    \begin{tabular}{c|ccc|cc}
    \toprule
      \multirow{2}{*}{Method}& \multicolumn{3}{c|}{Recall\%} & \multicolumn{2}{c}{mRecall\%} \\
      \cline{2-6}
       & Rel & Obj. & Pred. &  Obj. & Pred \\
      \hline
       IMP \cite{wu2023incremental} & 25.8 & 51.8 & 90.4 & 30.0 & 23.0 \\
       VGfM \cite{wu2023incremental}  & 28.3 & 53.3 & 90.7 & 31.6 & 24.4 \\
       3DSSG \cite{wu2023incremental} & 17.5 & 41.4 & 88.2 & 31.9 & 26.6 \\
       SGFN \cite{wu2023incremental} & 31.4 & 56.7 & 89.6 & 38.3 & 30.5 \\
       JointSSG \cite{wu2023incremental} & 34.1 & 58.1 & 89.9 & 43.0 & 33.3 \\
       \hline
       IMP  & 25.4 & 47.7 & 89.9 & 35.8 & 19.9 \\
       VGfM  & 27.4 & 50.1 & \cellcolor{best} \textbf{90.6} & 33.6 & 22.3 \\
       3DSSG & 13.0 & 35.6 & 86.8 & 26.0 & 24.3 \\
       SGFN & 29.9 & 52.2 & 89.5 & 35.9 & 27.1 \\
       JointSSG & 32.9 & 54.5 & 88.7 & 45.4 & 35.2 \\
       JointSSG$\ddagger$ & 34.3 & 56.6 & 86.5 & 52.9 & 36.2 \\
       \hline
       Ours$\dagger$ & \cellcolor{secbest} 38.6 & \cellcolor{secbest} 59.3 & 89.1 & \cellcolor{secbest} 57.4 & \cellcolor{secbest} 36.8 \\
       Ours & \cellcolor{best} \textbf{40.5} & \cellcolor{best} \textbf{61.8} & \cellcolor{secbest}90.4 & \cellcolor{best} \textbf{60.5} & \cellcolor{best} \textbf{39.2}\\
    \bottomrule     
    \end{tabular}
    \caption{Comparison with state-of-the-art methods on the 3RScan dataset with 20 object classes and 8 predicate classes. The top group of results are reported in JointSSG \cite{wu2023incremental}. The middle group of results are reproduced via 3DSSG GitHub repository. JointSSG$\ddagger$ refers to JointSSG but with DINOv2 multi-view image features. Ours$\dagger$ uses ResNet18 multi-view image features; Ours uses the DINOv2 multi-view image features instead. The \colorbox{best}{\textbf{Best}} and \colorbox{secbest}{Second Best} results are highlighted, respectively.}
    \label{tab:main}
    \vspace{-0.1in}
\end{table}

\begin{figure*}[!t]
  \centering
  \includegraphics[width=0.9\textwidth]{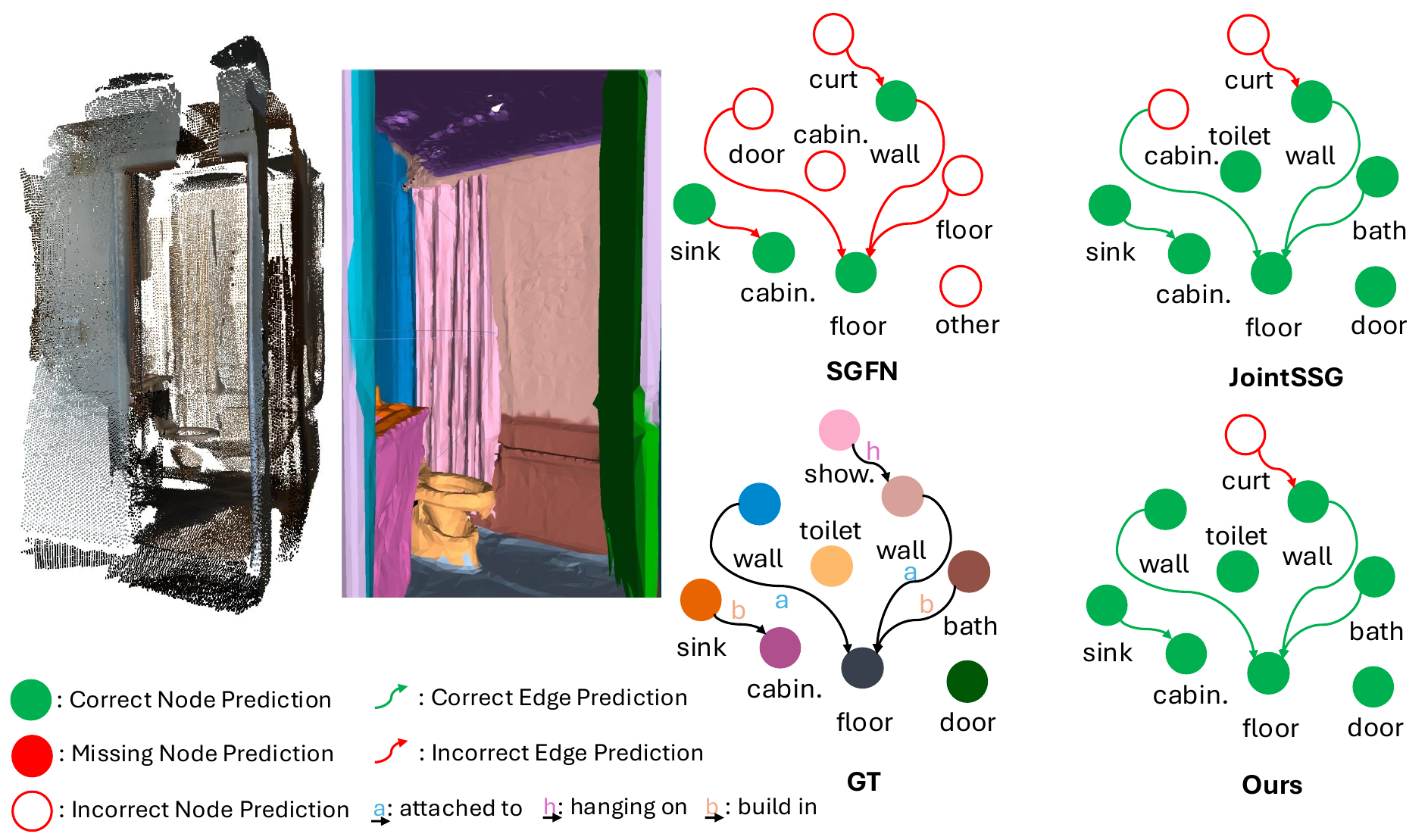}
  \vspace{-0.1in}
  \caption{The qualitative results on 3RScan dataset of previous works and our proposed framework on scan 4d3d82b0. SGFN also fails to classify majority of non-background object classes except the sink and cabinet. It also fails to classify all predicate classes. JointSSG performs better but misclassifies the shower curtain as a curtain and the ambiguous blue wall as a cabinet. Our method correctly classifies the wall but misidentifies the shower curtain as a curtain. }
  \label{fig:qualitative1}
    \vspace{-0.1in}
\end{figure*}

\begin{figure*}[!t]
  \centering
  \includegraphics[width=0.85\textwidth]{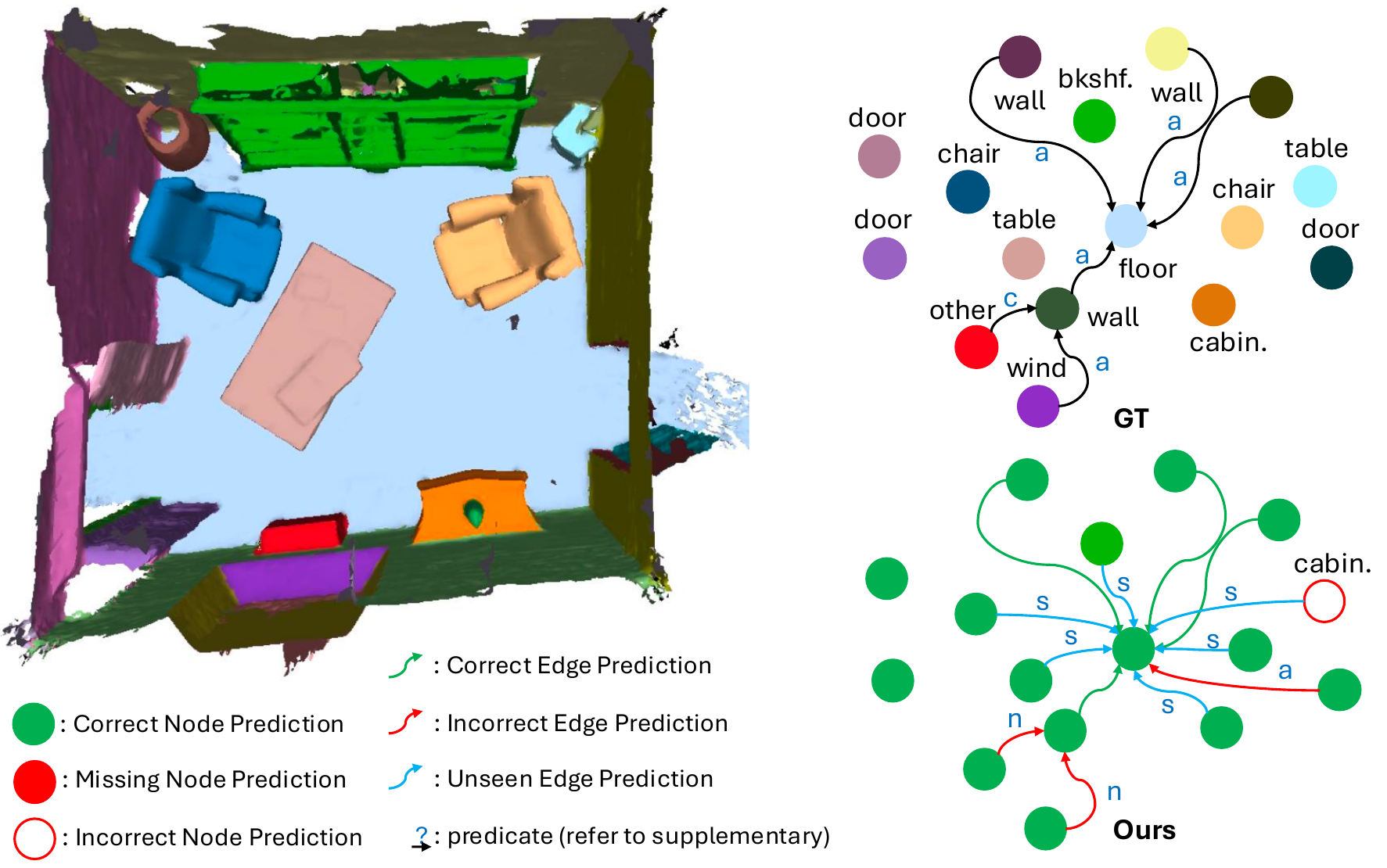}
  \caption{The qualitative results on 3RScan dataset of our proposed framework on the entire scene of scan 43b8cae1. Our method can correctly classify most objects in the scene except for the light blue dress table at the top right of the scan. Our method can also correctly classify most predicates.}
  \label{fig:qualitative_full2}
    \vspace{-0.1in}
\end{figure*}

\subsection{Qualitative Results}
We show qualitative results comparing our method with other competitive methods in Fig.~\ref{fig:qualitative1}. Since the scene contains a large number of objects, we display only a subset with fewer objects and predicates for better readability. The correct prediction of the \( None \) predicate class is omitted in the visualization due to its prevalence in the dataset. 
As seen in Fig.~\ref{fig:qualitative1}, our method can distinguish the relatively more ambiguous wall which the other methods fail to distinguish. The shower curtain is tricky to classify since it is a subset of a curtain that is another class of objects present in the dataset. 
Understandably, all methods fail to correctly classify this object.
Both our method and JointSSG attain good performance on the predicate class compared to SGFN as both methods only misclassify the predicate between the curtain object and the wall.

We additionally evaluate on the entire scene for scan 43b8cae1 without comparison to other methods
. 
Our method can recognize rare classes present in the scene, which are more challenging to learn.
As seen in Fig.~\ref{fig:qualitative_full2}, our approach successfully classifies the rare class bookshelf instance denoted by \(bkshf\) which no other method managed to predict based on the results shown in Tab.~\ref{tab:perclass}. Our method also predicts multiple reasonable predicates of \( standing\ on\) between various objects and \( floor \). Additional qualitative results
can be found in the supplementary material.

\subsection{Ablation Studies}

We analyze the effects of our feature initialization, robust neighbor residuals, and confidence rescoring module in Tab.~\ref{tab:ablation}.  
Our confidence rescoring (CR) module significantly improves performance across all metrics, indicating that it effectively leverages statistical priors to address class imbalance issues as noted in~\cite{wu2023incremental}. Apart from predicate mRecall, incorporating our masked feature initialization further improves performance across all metrics. This outcome is expected since masking primarily aims to refine multi-view image features, which do not directly contribute to edge feature improvements.  
The combination of the masked feature initialization and the CR module act as a force multiplier since the CR module benefits from cleaner and more accurate neighboring node features. Additionally, our robust neighbor residual gate maintains comparable performance across most metrics except for predicate mRecall, which benefits from additional neighbor information to better handle class imbalance in predicate estimation.

We additionally study the effects of using our CR module during inference in Tab.~\ref{tab:ablation_cr}. 
The addition of the CR module with the prior knowledge of the dataset is akin to calculating statistics during transductive setting of inference 
instead of a true overfitting to the dataset. Interestingly, the removal of the CR module during inference does not significantly reduce performance. It even leads to a slight increase in predicate Recall and overall relationship triplet Recall. The slight increase in predicate Recall, coupled with a drastic decrease in predicate mRecall, suggests that removing the CR module benefits the predicate classes that appear frequently in the dataset, including the \( None \) class, with the trade off of decreasing performance on the tail classes. Nonetheless, the use of 
the CR module during evaluation 
benefits object Recall, object mRecall, and predicate mRecall.



\begin{table}[]
    \centering
    \scalebox{0.77}{
    \begin{tabular}{c|ccc|ccc|cc}
    \toprule
      \multirow{2}{*}{Method}& \multirow{2}{*}{Conf.}  & \multirow{2}{*}{Feat.} & \multirow{2}{*}{Neigh.} & \multicolumn{3}{c|}{Recall\%} & \multicolumn{2}{c}{mRecall\%} \\
      \cline{5-9}
      & & & & Rel. & Obj. & Pred. &  Obj. & Pred. \\
      \hline
       JointSSG  & $\times$ & $\times$ & $\times$ & 32.9 & 54.5 & 88.2 & 45.4 & 35.2 \\
       \hline
       + Conf. & $\checkmark$ & $\times$ & $\times$ & 37.9 & 57.2 & 89.6 & 54.1 & 36.3 \\
       + Feat. & $\checkmark$ & $\checkmark$ & $\times$ & \textbf{41.1} & 61.3 & \textbf{90.8} & \textbf{61.3} & 35.8 \\
       \hline
       Ours & $\checkmark$ & $\checkmark$ & $\checkmark$ & 40.5 & \textbf{61.8} & 90.4 & 60.5 & \textbf{39.2} \\
    \bottomrule     
    \end{tabular}}
    \caption{Ablation study on the 3RScan dataset. Conf. refers to the usage of summary statistic priors for confidence rescoring via the CR module. Feat. refers to the usage of DINOv2 image features with refinement from SAM semantic masks via MFI. Neigh. refers to the usage of RSN-GNN. The \textbf{Best} results are shown in bold.}
    \label{tab:ablation}
\end{table}

\begin{table}[]
    \centering
    \scalebox{0.77}{
    \begin{tabular}{c|c|ccc|cc}
    \toprule
      \multirow{2}{*}{Method} & \multirow{2}{*}{CR.} & \multicolumn{3}{|c|}{Recall\%} & \multicolumn{2}{c}{mRecall\%} \\
      \cline{3-7}
         &  & Rel. & Obj. & Pred. &  Obj. & Pred. \\
      \hline
       Ours & $\times$ & \textbf{41.1} & 61.1 & \textbf{90.6} & 60.2 & 35.7 \\
       Ours & $\checkmark$ & 40.5 & \textbf{61.8} & 90.4 & \textbf{60.5} & \textbf{39.2} \\
    \bottomrule     
    \end{tabular}}
    \caption{Ablation study on the 3RScan dataset. CR. refers to the usage of summary statistic priors for confidence rescoring via the CR module during inference. The \textbf{Best} results are shown in bold.} 
    \label{tab:ablation_cr}
\end{table}

\vspace{2mm}
\noindent \textbf{Limitations.} 
The effectiveness of initial feature aggregation relies on the quality of the segmentation mask. Although the CR module performs well when there is no significant domain shift, 
the statistical prior can negatively impact performance when applied to datasets with a drastically different training data distribution. Similarly, the prior becomes less useful on datasets with very limited data.
\section{Conclusion}
\label{sec:conclusion}
We introduce a straightforward yet effective approach for enhancing 3D semantic scene graph estimation through our MFI, RSN-GNN, and CR modules. Specifically, our masked feature initialization (MFI) enhances node features by filtering out background distractors and leveraging more than just part-based image features. The RSN-GNN module strengthens edge features by integrating highly activated neighboring features into the target node feature, improving robustness. Finally, we introduce a refinement module that explicitly refines predictions using statistical prior knowledge, further enhancing performance.

\section*{Acknowledgments}
This research work is supported by the Agency for Science, Technology and Research (A*STAR) under its MTC Programmatic Funds (Grant No. M23L7b0021), and the Tier 2 grant MOE-T2EP20124-0015 from the Singapore Ministry of Education. We also thank Shun-Cheng Wu for the helpful discussions on running the SLAM code and reproducing the results in the 3DSSG repository. 
{
    \small
    \bibliographystyle{ieeenat_fullname}
    \bibliography{main}
}
\clearpage

\appendix
\setcounter{page}{1}
\maketitlesupplementary

\section{Additional Experiments}

\subsection{Non-{\bfseries\itshape{None}} Relationship Quantitative Results}
\label{subsec:non-none}
As shown in Tab. \ref{tab:nonone}, we additionally evaluate on the same 3RScan dataset 
without the \(None \) class for predicate and relationship prediction. 
We 
use the same metrics of top-1 Recall and mRecall. Several works \cite{wald2020learning, wu2021scenegraphfusion, wu2023incremental} 
consider the \(None \) relationships 
crucial, while others \cite{gay2019visual} only consider annotated non-\(None \) relationships. 
The latter approach avoids penalizing non-existent relationships that might otherwise be hallucinated while also preventing the model from overfitting to the prevalent \( None \) relationships. Nonetheless, it serves as an effective mechanism to verify whether the strong performance of the model in the former setting is a result of overfitting to the \( None \) relationship.


Our method performs better compared to other methods with less overfitting. Specifically, our predicate estimation surpasses all other methods even without the \(None \) relationships. This suggests that our method robustly learns the other non-\( None \) relationships.

\begin{table}[h]
    \centering
    \begin{tabular}{c|ccc|cc}
    \toprule
      \multirow{2}{*}{Method}& \multicolumn{3}{c|}{Recall\%} & \multicolumn{2}{c}{mRecall\%} \\
      \cline{2-6}
       & Rel & Obj. & Pred. &  Obj. & Pred \\
      \hline
       IMP \cite{wu2023incremental}  & 18.3 & 51.8 & 19.3 & 30.0 & 23.0 \\
       VGfM \cite{wu2023incremental} & 20.8 & 53.3 & 22.1 & 31.6 & 24.4 \\
       3DSSG \cite{wu2023incremental} & 15.1 & 41.4 & 26.1 & 31.9 & 26.6 \\
       SGFN \cite{wu2023incremental} & 24.4 & 56.7 & 27.2 & 38.3 & 30.5 \\
       JointSSG \cite{wu2023incremental} & 25.5 & 58.1 & 27.3 & 43.0 & 33.3 \\
       \hline
       IMP  & 14.4 & 48.1 & 16.8 & 35.8 & 19.9 \\
       VGfM  & 17.0 & 51.3 & 19.8 & 33.6 & 22.3 \\
       3DSSG & 11.9 & 36.2 & 25.7 & 26.0 & 24.3 \\
       SGFN & 21.5 & 53.8 & 24.6 & 35.9 & 27.1 \\
       JointSSG & 23.1 & 55.1 & 26.6 & 45.4 & 35.2 \\
       JointSSG$\ddagger$ & 22.4 & 56.8 &\cellcolor{secbest} 26.9 & 52.9 & 36.2 \\
       \hline
       Ours$\dagger$ &\cellcolor{secbest} 24.2 &\cellcolor{secbest} 60.1 & 26.4 &\cellcolor{secbest} 57.4 & \cellcolor{secbest} 36.8 \\
       Ours & \cellcolor{best} \textbf{25.7} & \cellcolor{best} \textbf{61.1} & \cellcolor{best} \textbf{27.6} & \cellcolor{best} \textbf{60.5} & \cellcolor{best} \textbf{39.2}\\
    \bottomrule     
    \end{tabular}
    \caption{Comparison with state-of-the-art methods on the 3RScan dataset with 20 object classes and 8 predicate classes. The top group of results are reported in \cite{wu2023incremental}. The middle group of results are reproduced via 3DSSG GitHub repository. JointSSG$\ddagger$ refers to JointSSG but with DINOv2 multi-view image features. Ours$\dagger$ refers to our method but with ResNet50 multi-view image features; Ours uses the DINOv2 multi-view image features instead. The \colorbox{best}{\textbf{Best}} and \colorbox{secbest}{Second Best} results are highlighted, respectively.} 
    \label{tab:nonone}
\end{table}


\subsection{Quantitative Results on 160 Object and 26 Predicate Classes}
We show the results of the evaluation on 160 object and 26 predicate classes for the 3RScan dataset similar to prior works. Direct comparisons on these metrics are unfair, they rely on ground truth point clouds and instance segmentation masks we do not use. We leverage only multi-view images with predicted instance masks and depth. Our method underperforms against the other methods as expected due to the inferior quality of predicted point clouds compared to the ground truth. However, it still performs comparably to SGFN for object-related metrics. 

\begin{table}[h]
    \centering
    \begin{tabular}{c|ccc|cc}
    \toprule
      \multirow{2}{*}{Method}& \multicolumn{3}{c|}{Recall\%} & \multicolumn{2}{c}{mRecall\%} \\
      \cline{2-6}
       & Rel & Obj. & Pred. &  Obj. & Pred \\
      \hline
       SGFN(GT) \cite{wu2023incremental} & 64.7 & 36.9 & 48.4 & 16.2 & 14.4 \\
       JointSSG(GT) \cite{wu2023incremental} & 67.6 & 53.4 & 48.1 & 28.9 & 24.7 \\
       \hline
       Ours(Pred) & 57.7 & 32.1 & 10.9 & 19.6 & 2.3 \\
    \bottomrule    
    \end{tabular}
    \caption{Comparison with state-of-the-art methods on the 3RScan dataset with 160 object classes and 26 predicate classes. The top group of results are reported in \cite{wu2023incremental} which uses ground truth point clouds and instance segmentation masks. We leverage only multi-view images with predicted instance masks and depth} 
    \label{tab:full}
\end{table}

\subsection{More Qualitative Results}
\begin{table}[]
    \centering
    \begin{tabular}{c|c}
        a & attached to  \\
        b & build in \\
        c & connected to \\
        h & hanging on \\
        n & none \\
        p & part of \\
        s & standing on \\
        u & supported by
    \end{tabular}
    \caption{
    Nomenclature} for predicates in the diagrams of our main paper. 
    \vspace{-3mm}
    \label{tab:notation}
\end{table}


For brevity in the main paper diagrams, we refer the reader to Tab.~\ref{tab:notation} for the nomenclature of the predicate notation.

We show the entire scene for scan 4d3d82b0 utilizing a similar layout format to scan 43b8cae1 in Fig.~\ref{fig:qualitative_full2}. 
Furthermore, we show the partial scene for scan 43b8cae1 using the layout format of scan 4d3d82b0. 

\begin{figure*}[!t]
  \centering
  \includegraphics[width=0.9\textwidth]{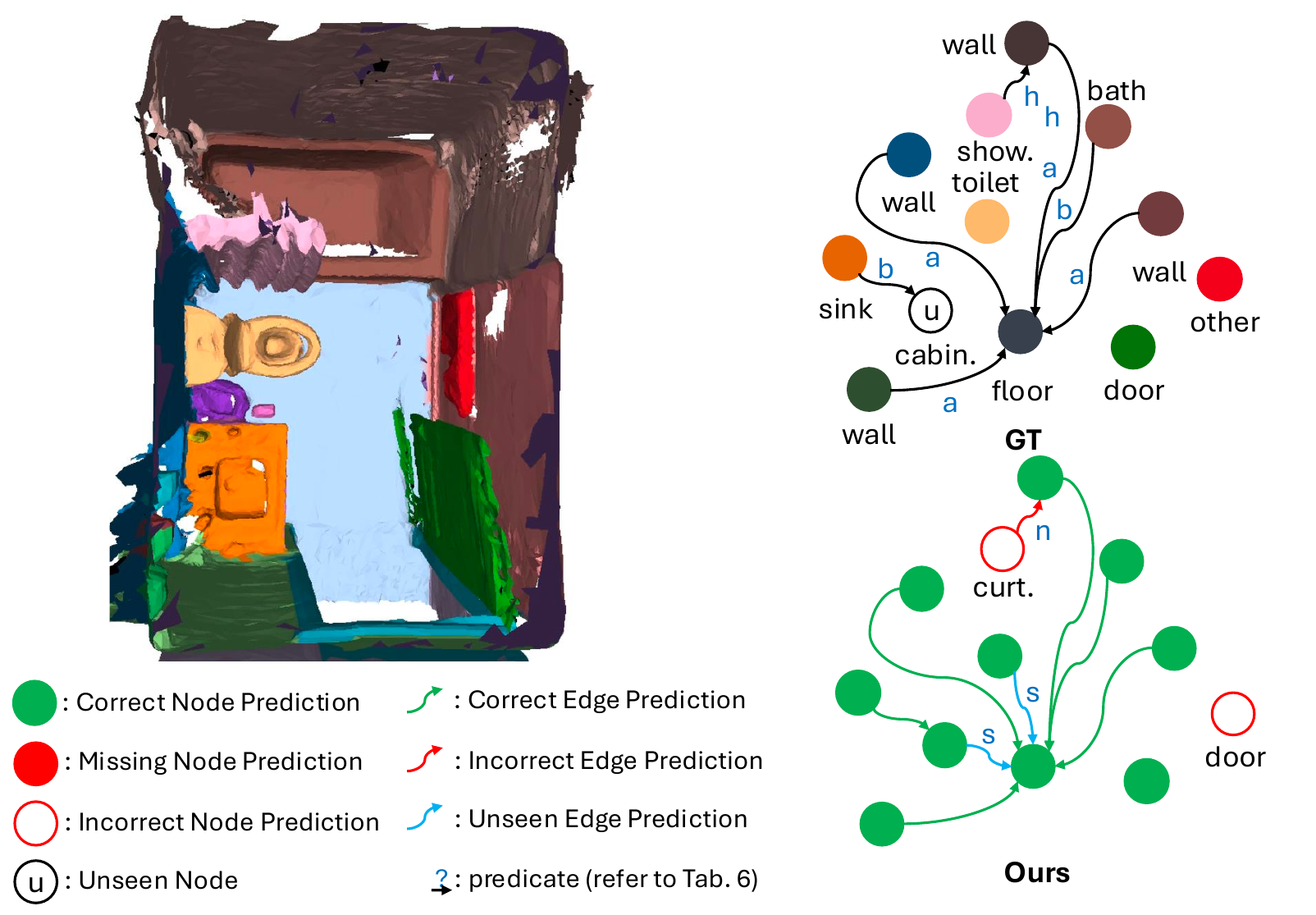}
  \vspace{-0.1in}
  \caption{The qualitative results on 3RScan dataset on the entire scene of scan 43b8cae1. Our method can correctly classify most objects in the scene except the misclassified pink shower curtain and other furniture depicted in red. Our method can reliably predict all predicate classes in the scene except for the predicate between the wall and the shower curtain.}
  \label{fig:qualitative_full1}
    \vspace{-0.1in}
\end{figure*}

In Fig.~\ref{fig:qualitative_full1}, in addition to the incorrect node prediction for the curtain in the partial scene, we observe that our model also mis-classifies another furniture object as a door. This mis-classification likely arises due to the thin structure of the object, which closely resembles that of a door.

\begin{figure*}[!t]
  \centering
  \includegraphics[width=0.9\textwidth]{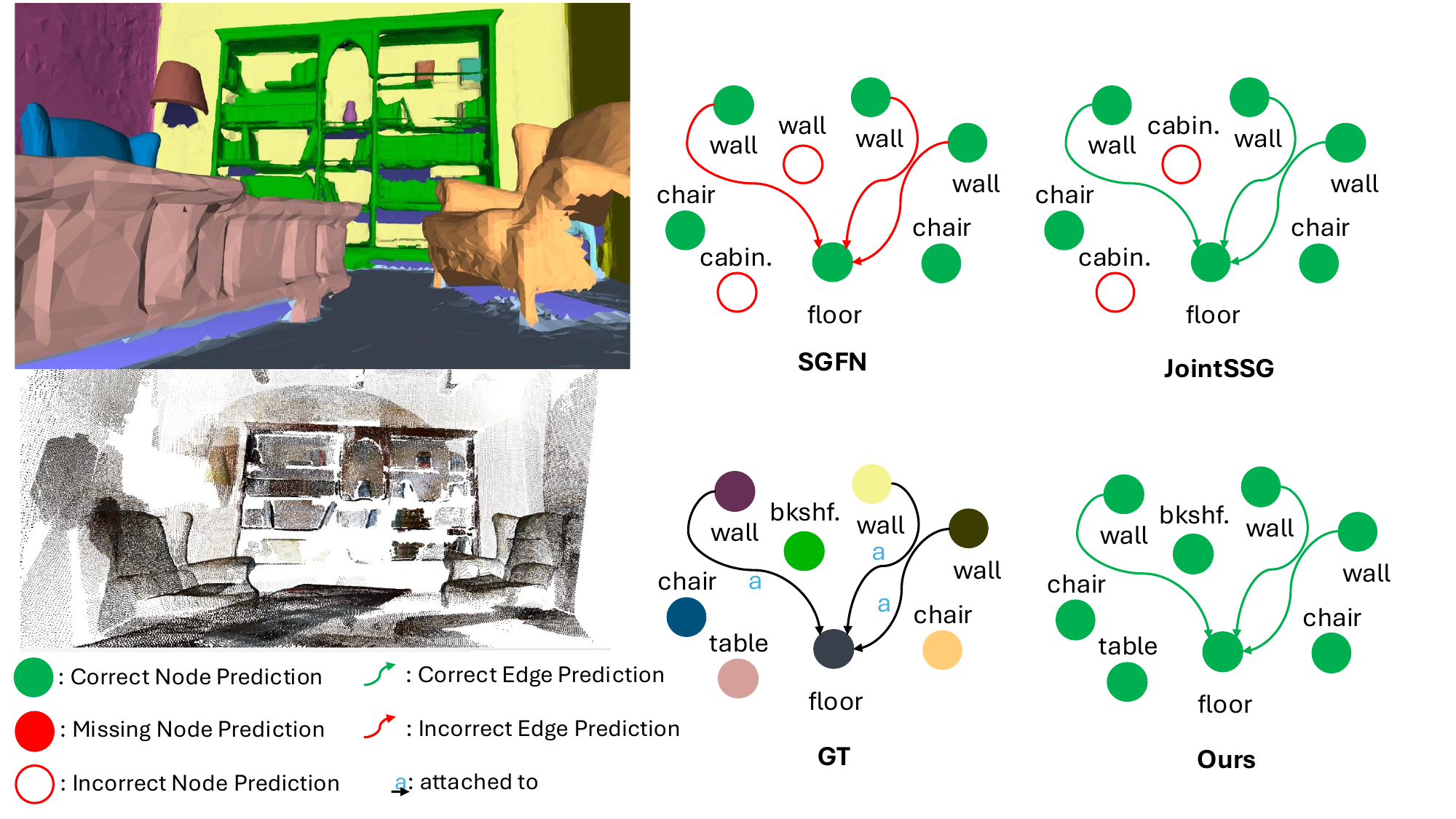}
  \vspace{-0.1in}
  \caption{The qualitative results on 3RScan dataset of previous works and our proposed framework on scan 43b8cae1. Our method can correctly classify all objects in the partial scene, including the green bookshelf and pink table that competing methods misclassify.}
  \label{fig:qualitative2}
    \vspace{-0.1in}
\end{figure*}

In Fig.~\ref{fig:qualitative2}, only our method successfully classifies the rare bookshelf class. Our method also correctly identifies an office table with a distinctive shape while other methods misclassify it as a cabinet. This demonstrates the superior ability of our method to recognize both rare object classes and uncommon shapes of common objects, such as tables.

\begin{figure*}[!t]
  \centering
  \includegraphics[width=0.9\textwidth]{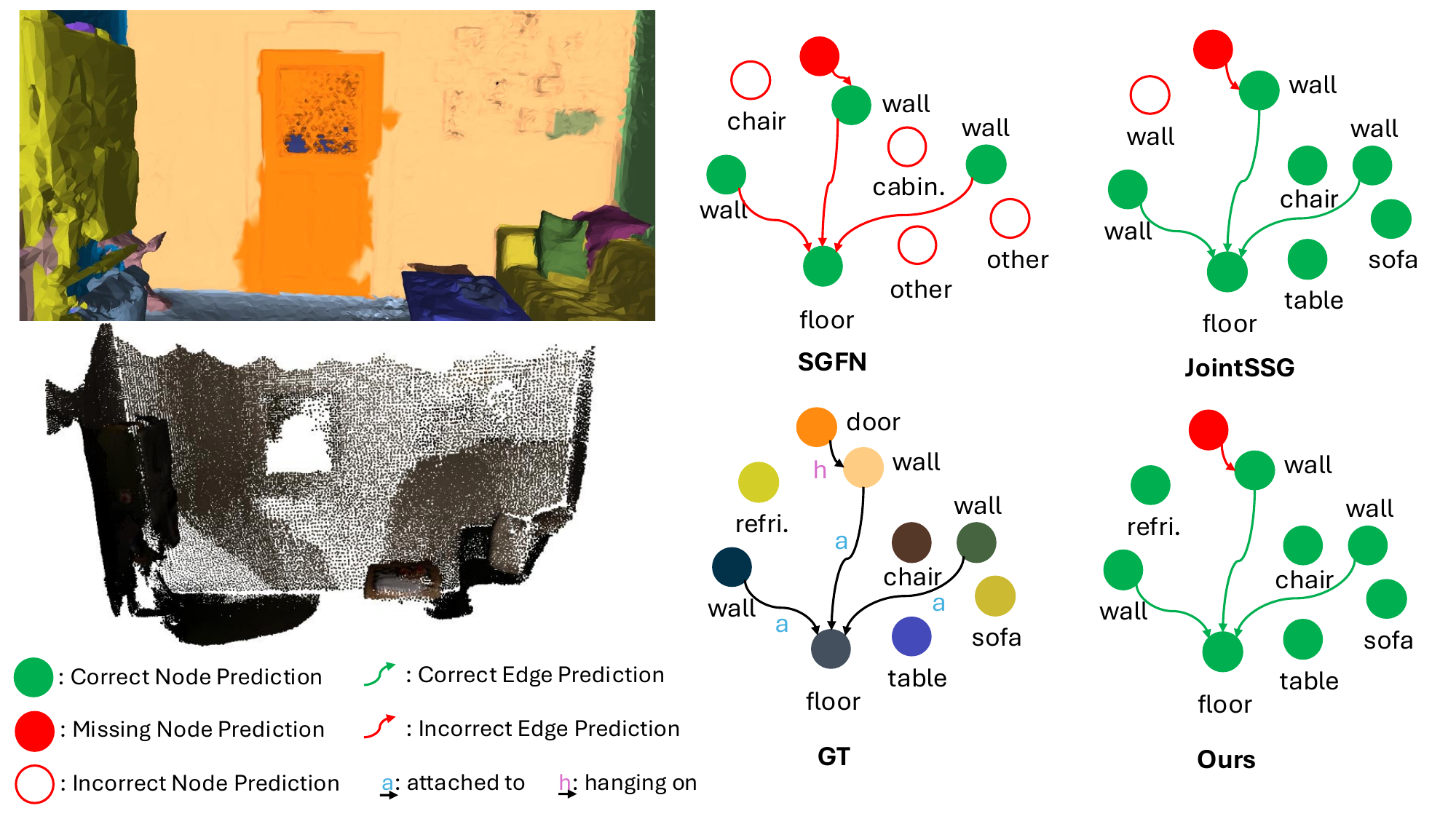}
  \vspace{-0.1in}
  \caption{The qualitative results on 3RScan dataset of previous works and our proposed framework on scan 4a9a43. SGFN fails to classify non-background object classes and all predicate classes. JointSSG performs better but misclassifies the refrigerator as a wall. The method also fails to detect the door. Our method correctly classifies the refrigerator but it also fails to detect the door.}
  \label{fig:qualitative}
    \vspace{-0.1in}
\end{figure*}
\begin{figure*}[!t]
  \centering
  \includegraphics[width=0.9\textwidth]{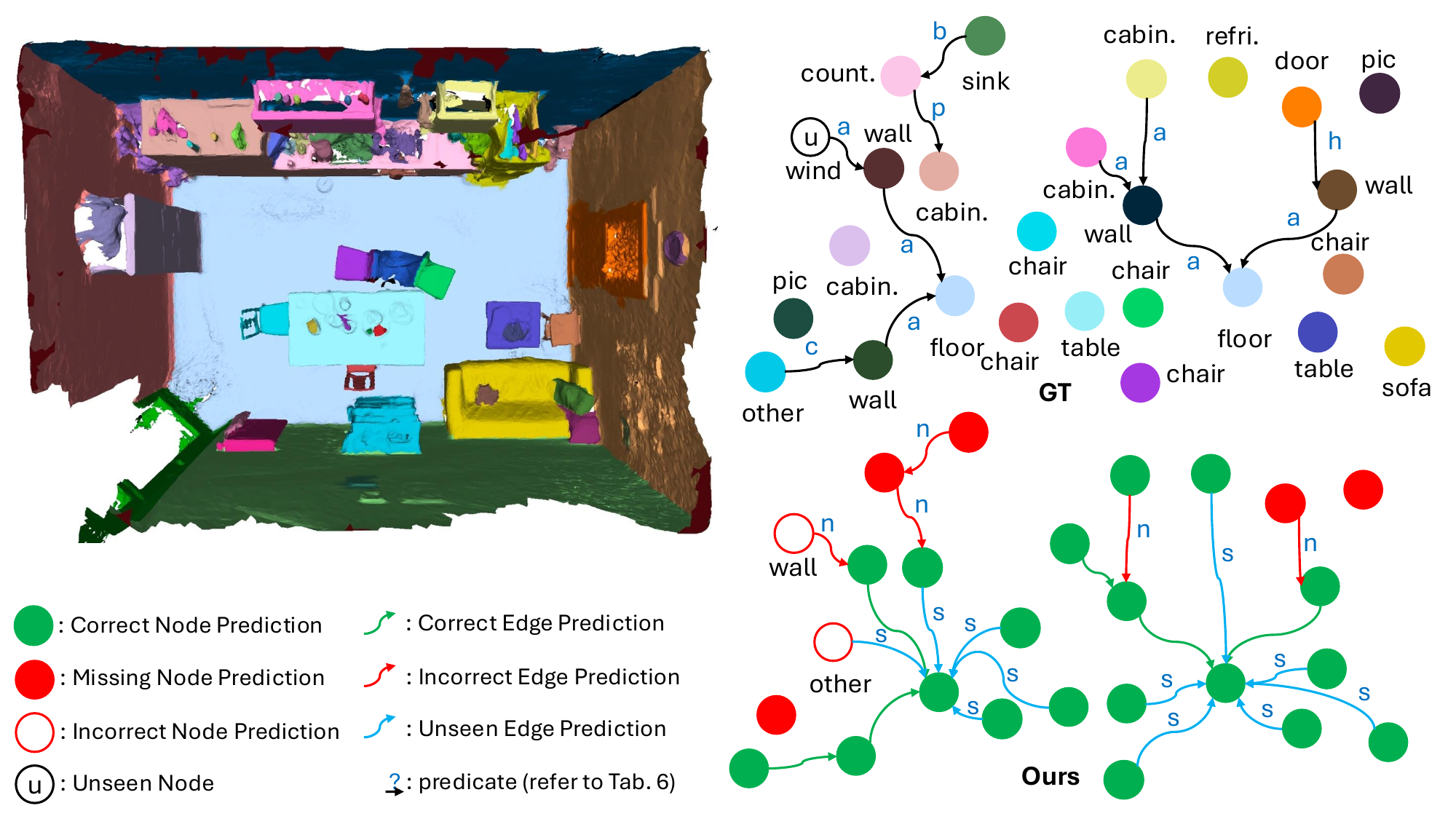}
  \vspace{-0.1in}
  \caption{The qualitative results on 3RScan dataset of our proposed framework on the entire scene of scan 43b8cae1. Our method can correctly classify most objects in the scene, though it may fail to predict the presence of some objects such as counter, picture, and sink. Our method can predict most predicate classes in the scene except for predicates with undetected neighboring objects.}
  \label{fig:qualitative_full}
    \vspace{-0.1in}
\end{figure*}

We provide qualitative results on one additional scan with the same format of partial scene in Fig.~\ref{fig:qualitative} and full scene in Fig.~\ref{fig:qualitative_full}, respectively.

In Fig.~\ref{fig:qualitative}, SGFN fails to classify non-background object classes and all predicate classes. Although JointSSG outperforms SGFN, it misclassifies the rare refrigerator class and fails to detect the orange door. Our method correctly classifies the refrigerator, but also misses the door. As a result, the predicate linking the door to the wall is misclassified by all methods.

In the full scene shown in Fig.~\ref{fig:qualitative_full}, our method correctly classifies most objects, but misses predictions on objects that are more difficult to discern. For example, counter, picture, and sink. Additionally, our method fails to classify the window, which is obscured beyond the bottom left of the scene. The light purple cabinet on the left is misclassified as other furniture, likely due to its atypical orientation compared to standard cabinets. 

Our method also fails to predict \(build\ in\) and \(part\ of\) predicates since it does not recognize the presence of the counter and sink. However, it successfully predicts the challenging \(connected\ to\) relationship between the dark green wall and the blue other furniture. Furthermore, it generates multiple plausible unseen predicate predictions for the \(standing\ on\) relationship between various objects and the floor.


\section{Analysis}
\subsection{Analysis on Logits}

\begin{figure*}[t]
  \centering
   \includegraphics[width=0.95\textwidth]{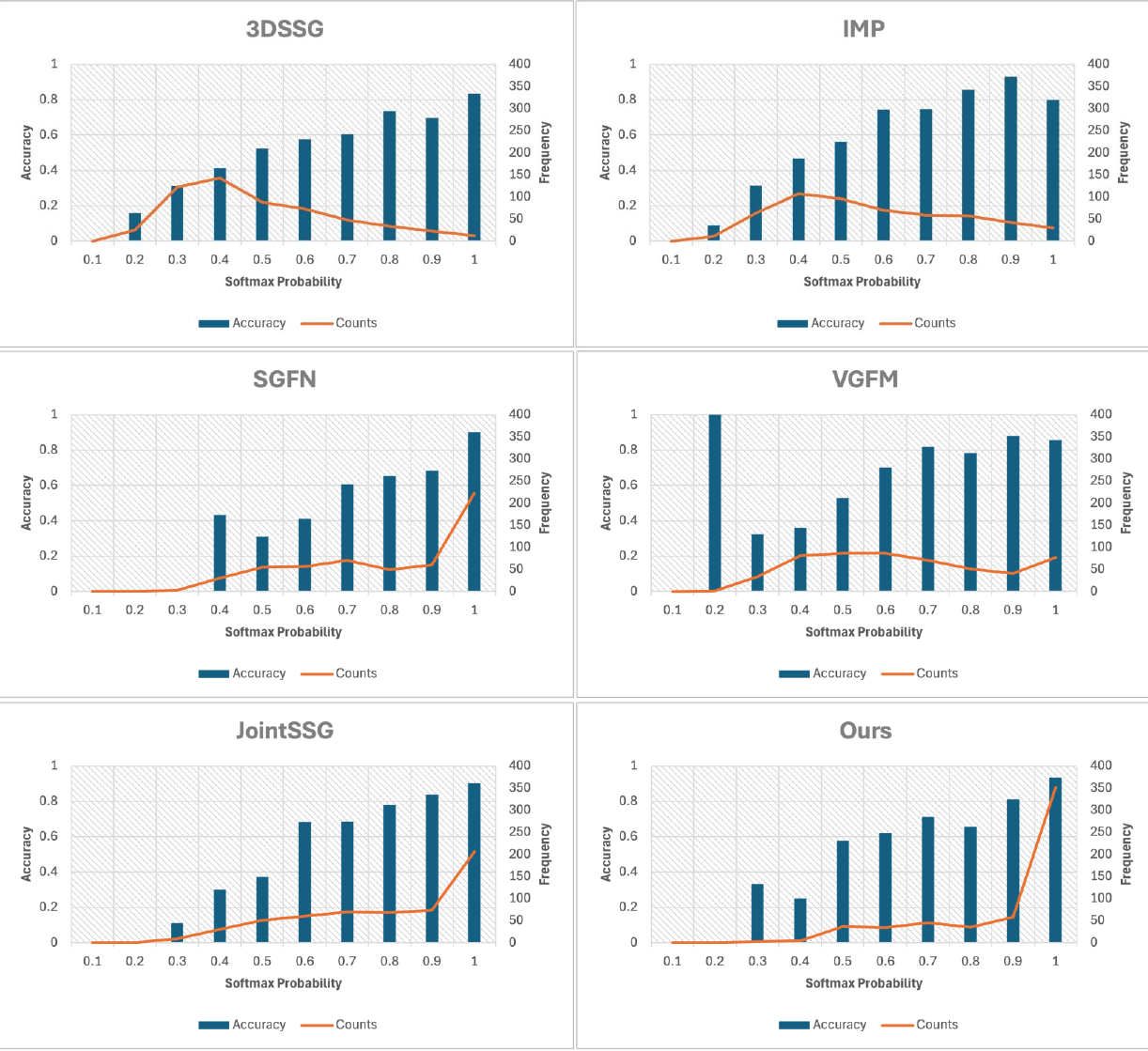}
   \vspace{-0.1in}

   \caption{Histogram of accuracy of object predictions with pseudo confidence bins for previous methods. More of our object predictions falls within the bins to the right as shown by the line graph suggesting more object predictions achieves higher pseudo confidence levels. Our method obtains higher accuracy at higher pseudo confidence levels (softmax probability) as shown by the bar graph. }
   \label{fig:accuracy}
   \vspace{-0.1in}
\end{figure*}

We analyze the impact of softmax logits on node estimation accuracy in Fig.~\ref{fig:accuracy} using the 3RScan test set. To assess this, we bin softmax probability predictions into \(0.1\) intervals and evaluate both the accuracy (represented by the bar chart) and the frequency of object predictions (represented by the line curve) within each bin. Each bin includes instances within the range \([x-0.1, x)\), where \(x\) corresponds to the labeled bin on the x-axis. The softmax probabilities serve as a pseudo-confidence measure for each prediction.

For improved model performance, it is desirable to minimize the number of low-confidence predictions as indicated by the line curve for 3DSSG in Fig.~\ref{fig:accuracy}. This is because lower-confidence predictions generally correspond to lower accuracy. 3DSSG exhibits a high frequency of low-confidence predictions with softmax probabilities below 0.5 and fewer high-confidence predictions. Similarly, IMP and VGFM also suffer from this issue. The unusually high performance of VGFM in the \([0.1, 0.2)\) bin is likely an outlier and can be ignored.

As evidenced by the rightward shift in the line curve, SGFN and JointSSG reduce the number of low-confidence predictions. Our method exhibits an even more pronounced shift, culminating in a sharp peak of nearly 400 object instances in the \([0.9, 1.0]\) bin. Additionally, our method achieves the highest accuracy in this bin, indicating that the majority of high-confidence predictions are correct. Combined with the insights from Fig.~\ref{fig:logits}, this suggests that the CR module plays a crucial role in increasing high-confidence predictions by enhancing the predicted node probability for the correct class.

\subsection{Analysis on CR Components}
To investigate the impact of the components of the CR module, we conduct experiments where we fix the confidence score \(\alpha\).

As shown in Tab.~\ref{tab:alpha}, CR fails without the confidence score \( \alpha \) modulating the input of the statistical prior, resulting in worse performance than our reproduced baseline for most metrics except predicate recall. For predicate recall, as established in ~\ref{subsec:non-none}, the models might overfit to the \( None \) relationship. As the mRecall for predicate estimation does not correspondingly improve with the predicate recall, this suggests that the CR module with a fixed \( \alpha \) does not help the model robustly learn the other non-\(None\) relationships. The adaptive value of \( \alpha \) is crucial to attaining good performance on the task.

\begin{table}[]
     \centering
    \begin{tabular}{c|ccc|cc}
    \toprule
      \multirow{2}{*}{Method}& \multicolumn{3}{c|}{Recall\%} & \multicolumn{2}{c}{mRecall\%} \\
      \cline{2-6}
       & Rel & Obj. & Pred. &  Obj. & Pred \\
      \hline
        Ours & 40.5 &  61.8 & 90.4 &  60.5 & 39.2\\  
        \hline
        \( \alpha=0.8 \) & 37.8 & 58.1 & 91.6 & 53.5 & 27.8 \\
        \( \alpha=0.2 \) & 31.0 & 53.0 & 90.5 & 46.5 & 26.2 \\
    \bottomrule 
    \end{tabular}
    \caption{Experiment on our method with a fixed confidence score threshold.}
    \label{tab:alpha}
\end{table}

\subsection{Analysis on CR Generalization Ability}
The dependency of CR on fixed co-occurrence statistics from the training data is a limitation of our approach. To evaluate robustness to distribution shifts, we conduct a simulated experiment by removing frequently occurring relationships from the dataset. 

Tab.~\ref{tab:shift} shows that there is a significant decrease in performance for predicate and relationship estimation while object estimation performance metrics only decreased slightly. As the \( None \) relationship is included in the top relationships to be removed, we attached the experiment without the \( None \) relationship for reference. The decrease in performance for predicate and relationship estimation compared to the experiment referenced above is less significant, which suggests some level of robustness of our method to distribution shifts.

For a broader generalization, adaptive co-occurrence statistics from visual foundation models could be explored, but this lies beyond the scope of our current closed-set setting.

\begin{table}[]
     \centering
    \begin{tabular}{c|ccc|cc}
    \toprule
      \multirow{2}{*}{Method}& \multicolumn{3}{c|}{Recall\%} & \multicolumn{2}{c}{mRecall\%} \\
      \cline{2-6}
       & Rel & Obj. & Pred. &  Obj. & Pred \\
      \hline
        Ours & 40.5 &  61.8 & 90.4 &  60.5 & 39.2\\  
        Ours (w/o \( None\)) & 25.7 & 61.1 & 27.6 &  60.5 & 39.2\\ 
        \hline
        Remove Top 20\% & 24.5 & 61.1 & 27.0 & 59.5 & 26.6 \\
       Remove Top 50\% & 18.9 & 60.1 & 20.6 & 56.5 & 19.1 \\
    \bottomrule 
    \end{tabular}
    \caption{Simulated experiment on our method's robustness to distribution shifts. We remove the most frequently occurring relationships according to the arbitrarily determined percentage in the first column of the table. }
    \label{tab:shift}
\end{table}

\end{document}